\documentclass[runningheads]{llncs}

\usepackage[T1]{fontenc}
\usepackage{graphicx}
\usepackage{booktabs}
\usepackage{multirow}
\usepackage{colortbl}
\usepackage[pagebackref,breaklinks,colorlinks]{hyperref}

\usepackage{wrapfig}
\usepackage{orcidlink}
\usepackage{xr-hyper}
\usepackage{algorithm}
\usepackage{algpseudocode}
\usepackage{amsmath,bm}
\usepackage{bbm}
\usepackage{amsfonts}
\usepackage{caption}
\usepackage{subfig}

\usepackage{graphicx}
\usepackage{color} 

\definecolor{Gray}{gray}{0.9}
\definecolor{Better}{rgb}{0.18, 0.407, 0.266}
\definecolor{Worse}{rgb}{0.35, 0.35, 0.35}
\definecolor{drakgreen}{rgb}{0.38, 0.67, 0.38}
\definecolor{drakpurple}{rgb}{0.38, 0.27, 0.61}
\definecolor{granate}{rgb}{0.64, 0.16, 0.16}

\newcommand{\our}{\cellcolor{Gray}}
\newcommand{\best}[1]{\textbf{#1}}
\newcommand{\imp}[1]{$_{{\textbf{\textcolor{Better}{#1}}}}$}
\newcommand{\wor}[1]{$_{{\textbf{\textcolor{Worse}{#1}}}}$}
\newcommand{\nc}[1]{ \textcolor{granate}{#1}}

\usepackage[euler]{textgreek}
\usepackage{arydshln}
\usepackage{marvosym}

\newcommand{\ww}{\mathbf{w}}

\newcommand{\pp}{\mathbf{p}}                     
\newcommand{\xx}{\mathbf{x}}                     
\newcommand{\W}{\mathbf{W}}                     
\newcommand{\yy}{\mathbf{y}} 
\newcommand{\kmarg}{\mathbf{m}}                
\newcommand{\real}{\mathbb{R}}                   
\newcommand{\vv}{\mathbf{v}}                     
\newcommand{\ttt}{\mathbf{t}}                    
\newcommand{\temp}{\tau}                         
\newcommand{\domain}[1]{$\mathcal{D}_\text{#1}$} 

\begin{document}

\title{Trustworthy Few-Shot Transfer of Medical \\ VLMs through Split Conformal Prediction}
\titlerunning{Trustworthy Few-Shot Transfer of Medical VLMs}

\author{
Julio Silva-Rodríguez\textsuperscript{\Letter}
\and
Ismail {Ben Ayed}
\and
Jose Dolz
}

\authorrunning{J.~Silva-Rodríguez et al.}

\institute{
ÉTS Montréal \\ 
\Letter {\tt \small \email{julio-jose.silva-rodriguez@etsmtl.ca}} 
}

\index{Silva-Rodríguez, Julio}
\index{Ben Ayed, Ismail}
\index{Dolz, Jose}

\maketitle          

\begin{abstract}

Medical vision-language models (VLMs) have demonstrated unprecedented transfer capabilities and are being increasingly adopted for data-efficient image classification. Despite its growing popularity, its reliability aspect remains largely unexplored. This work explores the split conformal prediction (SCP) framework to provide trustworthiness guarantees when transferring such models based on a small labeled calibration set. Despite its potential, the generalist nature of the VLMs' pre-training could negatively affect the properties of the predicted conformal sets for specific tasks. While common practice in transfer learning for discriminative purposes involves an adaptation stage, we observe that deploying such a solution for conformal purposes is suboptimal since adapting the model using the available calibration data breaks the rigid exchangeability assumptions for test data in SCP. To address this issue, we propose \textit{transductive split conformal adaptation} (SCA-T), a novel pipeline for transfer learning on conformal scenarios, which performs an unsupervised transductive adaptation jointly on calibration and test data. We present comprehensive experiments utilizing medical VLMs across various image modalities, transfer tasks, and non-conformity scores. Our framework offers consistent gains in efficiency and conditional coverage compared to SCP, maintaining the same empirical guarantees. The code is publicly available: \url{https://github.com/jusiro/SCA-T} .

\end{abstract}

\section{Introduction}
\label{sec:intro}

Vision-language models (VLMs) pre-trained on extensive data sources, such as CLIP \cite{radford2021learning}, are witnessing widespread adoption in a plethora of computer vision problems \cite{clap24,shakeri2024few}, showcasing unprecedented transfer capabilities to downstream tasks. These models have rapidly gained traction in safety-critical scenarios, such as healthcare, where specialized medical VLMs have been recently introduced for radiology \cite{convirt,MedCLIP}, histology \cite{PLIP,CONCH}, or retina \cite{FLAIR}. In this context, ensuring their reliable deployment is paramount to minimize undesirable risks. The reliability of VLMs goes beyond its most explored discriminative aspect \cite{zhou2022coop,gao2021clip,clap24} and requires providing robust uncertainty estimates. Nevertheless, this crucial factor has often been overlooked in the literature, with only a handful of recent works mainly exploring the uncertainty of CLIP predictions from a calibration standpoint \cite{yoonc,sals,tuempirical}. Despite their popularity \cite{mehrtash2020confidence,murugesan2024class}, confidence calibration approaches lack theoretical guarantees of the actual model performance. In other words, these methods cannot estimate the most likely output and provide a verified probability of their predictive capabilities.

Conformal Prediction (CP) \cite{learning_ny_transduction,vovk_book} is a machine learning framework that provides model agnostic, and \textit{distribution-free}, finite-sample validity guarantees, ensuring reliability in predictive systems, and which has recently gained traction in modern neural networks \cite{raps,ding2024class}. In CP, the objective is to provide the user with prediction sets in which the probability of the correct label being included satisfies a user-specified coverage level, e.g., in the medical context, CP can ensure that the correct diagnosis (e.g., the presence of a given disease) is included in the predictive sets $95\%$ of the time. \textit{Split conformal prediction} (SCP) \cite{inductive_ci,vovk_book}, a variant of CP techniques, is gaining increasing attention for modeling uncertainties in vision black-box predictors \cite{raps,ding2024class}, i.e., models for which only access to their outputs is granted. SCP provides a practical scenario to incorporate such guarantees (marginalized over the entire test set) by leveraging a \textit{calibration set}, which is assumed to be exchangeable w.r.t. test data \cite{vovk_book}. Despite the promising performance of SCP in task-specific solutions, \cite{raps,aaai_medical_conf}, its deployment in modern specialized VLMs remains less explored. Indeed, VLMs are strong zero-shot predictors \cite{radford2021learning}, which can perform open-set black-box predictions on general concepts \cite{radford2021learning,Moor2023}. However, as recent evidence points out, its performance for such concepts highly relies on the pre-training concept 
and domain frequencies \cite{udandarao2024zeroshot,forgotten_domain_generalization}. Note that, while this limitation does not affect the coverage guarantees of SCP, as these are distribution-free, it can affect the efficiency of the final conformal solution and, consequently, its practical utility. Transfer learning in VLMs has been widely studied in recent literature to mitigate such domain gap, where black-box Adapters, e.g., linear probe strategies, have been proposed \cite{clap24,lp24} to adapt VLMs to downstream tasks by leveraging the embedding representations in the few-shot data regime. However, as we highlight in this work, this solution does not apply to the conformal prediction framework. In particular, adjusting these classifiers on the available small labeled calibration set might produce a covariate shift in the output probability distribution w.r.t. test data, breaking the exchangeability assumptions of CP. In this context, the theoretical guarantees of conformal prediction will not hold, producing unreliable predictive sets, as illustrated in Figure~\ref{fig:coverage_datasets}. This observation motivates the following question: \textit{Can we enhance the performance of medical VLMs in SCP settings while maintaining the same coverage guarantees on incoming test data?} 

\noindent The main contributions of this paper can be summarized as:
\begin{itemize}
    \item[$\circ$] We explore the SCP framework for medical VLMs, equipping them with trustworthiness guarantees in the popular few-shot scenario.
    \item[$\circ$] To address transfer learning in VLMs, we introduce \textit{transductive split conformal adaptation} (SCA-T). In this unsupervised framework, calibration and test data are jointly transferred to prevent breaking the exchangeability assumptions, thereby maintaining coverage guarantees.
    \item[$\circ$] We proposed a transductive solver built upon well-established knowledge on information maximization, enabling the introduction of a specific label-marginal regularization estimated from the calibration set. 
    \item[$\circ$] Comprehensive experiments on 9 public datasets and 3 modality-specialized medical VLMs demonstrate the benefits of SCA-T to enhance average efficiency and conditional coverage over SCP, with up to $+18\%$ gains.
\end{itemize}

\section{Related Work}
\label{sec:rw}

\noindent\textbf{\textit{VLMs transfer}} is typically carried out in the few-shot data regime, where Prompt Learning \cite{zhou2022coop} and black-box Adapters \cite{gao2021clip,lp24,clap24} are popular strategies. These have been recently assessed in medical VLMs \cite{shakeri2024few}, showing on-par performance, despite Adapters being more efficient, as they operate over embedding representations. Concretely, SoTA Adapters are text-informed linear probes with text ensembles \cite{lp24}, or constraint optimization objectives \cite{clap24}. However, these methods do not leverage the test data's underlying distribution; that is, they are not transductive, which is our focus. The recent SoTA for transductive black-box adaptation in VLMs is TransCLIP \cite{zanella2024boosting}, which combines well-established knowledge in unsupervised Gaussian Mixture Models with textual, zero-shot priors.

\noindent\textbf{\textit{Conformal prediction.}} Current trends on CP for black-box image classifiers follow the SCP framework \cite{raps}. These works have focused on providing non-conformity measures to create conformal sets, e.g., by improving adaptability. Some non-conformity scores are LAC \cite{lac}, which employs the raw probabilities; APS \cite{aps}, accumulating the total uncertainty of a target category by accumulating the sorted softmax scores; or RAPS \cite{raps}, which integrates penalties to promote smaller sets. Few works have explored SCP in medical image classification, e.g., by proposing ordinal non-conformity scores for grading tasks \cite{ordinalraps} or group-calibrated predictors \cite{aaai_medical_conf}, typically evaluated in specific applications. Note, however, that these models are intensively adjusted on a data corpus independent and identically distributed (i.i.d.) w.r.t. calibration/testing, an unrealistic scenario for pre-trained VLMs, which are our focus.

\noindent\textbf{\textit{Conformal prediction in VLMs.}} Some very recent works have explored CP in VLMs. Conf-OT \cite{confot25} proposes optimal transport to re-balance the zero-shot logits and address domain shift, and FCA \cite{fca25} deploys a full-conformal adaptation pipeline for both adapting and conformalizing the outputs. The solution presented in this work is more flexible than Conf-OT \cite{confot25}, since we assume access to the embedding representations. Despite being less flexible than FCA \cite{fca25}, our solution enables adaptation without scaling compute with task complexity, i.e., number of classes, a key limitation of full-conformal scenarios.

\section{Methods}
\label{sec:background}

\subsection{Zero-shot models}
\label{ssec:zero-shot}

\noindent\textbf{\textit{Contrastive VLMs}} are trained on large heterogeneous datasets to encode similar representations between paired image and text information. Their visual and text encoders project data points into an $\ell_{2}$-normalized shared embedding space, yielding the corresponding visual, ${\vv} \in \real^{D \times 1}$, and text, ${\ttt} \in \real^{D \times 1}$, embeddings. Given a pre-computed image feature and class-wise prototypes, $\W=(\ww_c)_{1 \leq c \leq C}$, with $\ww_c \in \real^{D \times 1}$, and $C$ the target classes, probabilities can be obtained as:
\begin{align}
\label{eq:probs}
    \phantom{,}p_c(\W,\vv)
    = \frac
    {\exp( \vv^\top \ww_{c} / \temp)}
    {\sum_{j=1}^{C} \exp( \vv^\top \ww_j / \temp)},
\end{align}

\noindent where $\vv^\top \ww$ is the dot product, and $\pp(\W,\vv)=(p_c(\W,\vv))_{1 \leq c \leq C}$ corresponds to the temperature-scaled predicted probabilities vector.

\noindent\textbf{\textit{Text-driven predictions.}} Contrastive VLMs enable zero-shot predictions, i.e., no need to explicitly learn $\W$, by embedding a textual description for each label into the weight matrix, i.e., $\ww_{c}=\ttt_{c}$. Generally, the zero-shot prototypes are the average of the $\ell_2$-normalized text embeddings using a set of $J$ text different templates, e.g., "\texttt{A photo of a [CLS]}.", such that $\ttt_{c}=\frac{1}{J}\sum_{j=1}^{J}\ttt_{cj}$.

\subsection{Conformal prediction}
\label{ssec:conformal}

\noindent\textbf{\textit{Preliminaries.}} Let us define an image classification task, where $(\xx,y)$ denotes random data points. Also, we denote a black-box deep network, $\pi(\cdot)$, which outputs probability assignments, $\pp=\pi(\xx)$, for a set of labels, $\mathcal{Y}=\{1, 2, ..., C\}$.

\noindent\textbf{\textit{Conformal prediction.}} In CP \cite{vovk_book}, the goal is to produce prediction sets that contain the true label with a user-specified probability. Formally, the aim is constructing a set-valued mapping $\mathcal{C}:\mathcal{X}\rightarrow 2^{C}$, such that:
\begin{align}
\label{eq:marginal}
    \mathcal{P}(Y\in \mathcal{C}(\xx)) \geq 1-\alpha,
\end{align}
\noindent where $\alpha \in (0, 1)$ denotes the desired error rate, and $\mathcal{C}(\xx) \subset \mathcal{Y}$ is the prediction set. Eq. \ref{eq:marginal} is known as the \textit{coverage guarantee} \cite{vovk_book}, and is \textit{marginal} over $\mathcal{XY}$. 

\noindent\textbf{\textit{Split conformal prediction.}} SCP \cite{inductive_ci} deploys coverage guarantees for any pre-trained black-box classifier \cite{Lei2018}. This framework uniquely requires a calibration set consisting of $N$ labeled data points, \domain{cal}$=\{(\xx_i,y_i)\}_{i=1}^{N}$, drawn from a data distribution which is at least exchangeable \cite{vovk_book} w.r.t. test data, \domain{test}$=\{(\xx_i)\}_{m=N+1}^{N+M}$. The SCP procedure is as follows: \textit{i}) \textit{First} a non-conformity score $s_{i}=\mathcal{S}(\pi(\xx_i),y_i)$ is defined, where $s_{i}$ is a degree of uncertainty; \textit{ii}) \textit{Second}, the 1-$\alpha$ quantile of the non-conformity score is determined from calibration data, which will serve as a confidence threshold to satisfy a given coverage:
\begin{align}
\label{eq:threshold}
    \hat{s} = \text{inf}
    \biggl[
    s : 
    \frac{ |i\in \{1,...,N\}: s_{i} \leq s| } { N }
    \geq 
    \frac{ \lceil (N+1)(1-\alpha) \rceil } { N }  
    \biggr];
\end{align}

\noindent \textit{iii}) \textit{Third}, non-conformity scores of each test data point and label are evaluated. The output sets are composed of the labels whose measure falls within $\hat{s}$:
\begin{align}
\label{eq:inference}
    \mathcal{C}(\xx) = \{ y \in \mathcal{Y} : \mathcal{S}(\pi(\xx),y) \leq \hat{s} \}.
\end{align}

\subsection{Transductive Split Conformal Adaptation (SCA-T)}
\label{ssec:setting}

\noindent\textbf{\textit{Transfer learning.}} We aim to enhance conformal sets of pre-trained VLMs. A straightforward solution would involve leveraging \domain{cal} supervision to adjust a linear probe, following relevant literature \cite{gao2021clip,clap24}. Nonetheless, as shown in Fig. \ref{fig:coverage_datasets} (Adapt+SCP), modeling the output distribution using these labels would lead to a covariate shift, resulting in unsatisfactory coverage guarantees on the test data. Hence, we propose using an unsupervised learning objective, not relying directly on calibration labels. Additionally, such optimization is transductive, i.e., it is performed on both the joint calibration and test data, ensuring uniform transfer and reducing potential distribution shifts between the two subsets.

\noindent\textbf{\textit{Transductive solver.}} Let us denote a calibration dataset composed of labeled embedding representations, $\{(\vv_i,y_i)\}_{i=1}^{N}$, and a set of exchangeable test data points, $\{(\vv_i)\}_{i=N+1}^{N+M}$. The objective is to learn a label assignment on the joint embedding dataset, $\{(\vv_i)\}_{i=1}^{N+M}$, by learning a set of optimal class prototypes, $p_c(\W^*,\vv_i)$, in an unsupervised manner. To solve this, we can leverage transductive objectives based on mutual information maximization, e.g., TIM \cite{tim}:
\begin{align}
\label{eq:tim}
    \phantom{.}\min_{\mathbf{W}} \ \ \ \frac{\lambda}{(N+M)} \sum_{i=1}^{N+M} \sum_{c=1}^C p_c(\W,\vv_i) \log (p_c(\W,\vv_i)) - \sum_{c=1}^C \hat{\kmarg} \log (\hat{\kmarg}) \,
\end{align}

\noindent where $\hat{\kmarg}=\frac{1}{N+M}\sum_{i=1}^{N+M} \pp(\W,\vv_i))$ is the estimated label-marginal distribution, and $\lambda$ is a blending hyper-parameter. The first term is the sample-conditional entropy, whose minimization aims to push the decision boundaries from dense regions. The second term refers to the label-marginal Shannon entropy, whose maximization encourages the marginal distribution of labels to be uniform, acting as a regularizer to avoid non-trivial solutions. The overall optimization objective in Eq. \ref{eq:tim} can be solved, for example, by gradient descent optimizers.

One can notice that the regularizer in \textit{TIM is biased toward uniform label distributions and does not generalize to imbalanced data regimes}, which are natural in medical tasks. In SCP, it is worth noting that we have access to the properties of the target data distribution, which can be inferred from calibration data. In particular, the marginal distribution is known and can be calculated from the one-hot encoded labels, $\yy_i$, such that $\kmarg=\frac{1}{N}\sum_{i=1}^N \yy_i$. Hence, we propose a mutual information solver that can absorb this information. Concretely, we replace the Shannon entropy term with a more flexible Kullback–Leibler divergence, which regularizes the overall solution toward the observed priors:
\begin{align}
\label{eq:tim_kl}
    \phantom{.}\min_{\mathbf{W}} \ \ \ \frac{\lambda}{(N+M)} \sum_{i=1}^{N+M} \sum_{c=1}^C p_c(\W,\vv_i) \log (p_c(\W,\vv_i)) + \sum_{c=1}^C \kmarg \log \frac{\kmarg}{\hat{\kmarg}} \,
\end{align}

\noindent\textbf{\textit{Conformal sets.}} The final step is producing the output sets using the new classifier's scores, $\pp(\W^*,\vv)$. First, calibration, $\{({\pp(\W^*,\vv_i)}^\top,y_i)\}_{i=1}^{N}$, and test, $\{(\pp(\W^*,\vv_i))\}_{i=N+1}^{N+M}$, data is separated again. Second, vanilla SCP is followed as in Section \ref{ssec:conformal}, but using the updated classifier (from Eq. \ref{eq:tim_kl}): \textit{i}) non-conformity scores, $s_{i}=\mathcal{S}(\pp(\W^*,\vv_i),y_i)$, are generated; \textit{ii}) the $1-\alpha$ quantile is search as in Eq. \ref{eq:threshold}, and \textit{iii}) conformal sets are generated on test data following Eq. \ref{eq:inference}.

\section{Experiments}
\label{sec:experiments}

\subsection{Setup}

\noindent\textbf{\textit{Medical VLMs.}} Three medical modalities are explored. \textbf{Histology}: CONCH \cite{CONCH} is used as a histology-specialized model, using a ViT-B/16 visual backbone. \textbf{Ophtalmology:} FLAIR \cite{FLAIR} is selected. \textbf{CXR}: CONVIRT \cite{convirt} pre-trained on MIMIC \cite{mimic} is used. FLAIR and CONVIRT follow the same architectural design, composed of BioClinicalBERT \cite{bioclinicalbert} and ResNet-50 encoders.

\noindent\textbf{\textit{Downstream tasks.}} Balanced and imbalanced datasets, for grading and fine-grained classification, are employed. \textbf{Histology}: colon in NCT-CRC \cite{kather2018100}, prostate Gleason grading in SICAPv2 \cite{silva2020going}, and SkinCancer \cite{kriegsmann2022deep} are evaluated. \textbf{Ophtalmology}: diabetic retinopathy grading using MESSIDOR \cite{messidor}, myopic maculopathy staging in MMAC \cite{mmac}, and disease classification in FIVES \cite{fives}, are assessed. \textbf{CXR}: five CheXpert \cite{irvin2019chexpert} categories from \cite{MedCLIP}, fine-grained categories from NIH-LT \cite{nih,nih_lt}, and multi-class pneumonia classification in COVID \cite{covid1,covid2}.

\noindent\textbf{\textit{Transfer learning.}} Our transductive solver is adjusted using the joint calibration and test data. The class weights are initialized using the text prototypes. Then, they are updated to minimize Eq. \ref{eq:tim_kl} via full-batch gradient descent during 100 iterations. ADAM is the optimizer, with a base learning rate of 0.01 and a cosine decay scheduler. Finally, $\lambda$ is set to 1.0. \textit{Note that all hyper-parameters are fixed across all tasks and data regimes.}

\noindent\textbf{\textit{Baselines.}} Relevant black-box transductive baselines are included in our evaluation. First, the original TIM \cite{tim} is used. We select the gradient-based version, with similar training details as our method, maintaining the relative weight of the sample conditional entropy to 1, as it yields best performance. Second, we leverage the recent SoTA method for transduction using VLMs, i.e., TransCLIP~\cite{zanella2024boosting}.

\noindent\textbf{\textit{Conformal prediction:}} Three popular non-conformity scores for image classification are assessed: LAC \cite{lac}, APS \cite{aps}, and RAPS \cite{raps}, with RAPS's hyper-parameters set to $k_{\text{reg}}=1$ and $\lambda=0.001$, which provided stable performance in \cite{raps}. Furthermore, prediction sets at error rates of $\alpha \in \{0.1, 0.05\}$ are generated. 

\noindent\textbf{\textit{Experimental protocol.}} \textbf{Calibration data:} $N=C\times K$ data points are sampled from training data following the label-marginal distribution of the downstream tasks, where $K=16$ indicates the number of samples per class in the few-shot literature \cite{shakeri2024few}. \textbf{Conformal strategies:} \textit{i}) \textbf{SCP}, the standard split conformal prediction studied in vision classifiers \cite{raps,ding2024class} which operates over zero-shot predictions; \textit{ii}) \textbf{Adapt+SCP}, which first adjusts a linear classifier and then follows the SCP over the new predictions, using the same calibration set for adaptation and calibration; and \textit{iii}) \textbf{SCA-T}, the proposed transductive strategy that performs the transfer on the joint calibration and test data before conformal prediction. \textbf{Metrics} employed in conformal prediction literature \cite{fca25} are computed: balance classwise accuracy (ACA), coverage (“Cov.”), average set size (“Size”), and class-conditioned coverage gap (“CCV”) \cite{ding2024class}. Regarding computational efficiency, we assessed peak GPU memory usage in gigabytes (“GPU”) and inference time in seconds (“T”). The experiments are repeated 100 times with different sampling seeds.

\begin{table}[h!]
\caption{\textbf{Conformal prediction results}, averaged across modalities and tasks.}
\label{main_results}
\centering
\begin{tabular}{llccccccc}
\toprule
\multicolumn{2}{c}{\multirow{2}{*}{Method}} & & \multicolumn{3}{c}{$\alpha=0.10$} & \multicolumn{3}{c}{$\alpha=0.05$}   \\ \cmidrule(lr){4-6} \cmidrule(lr){7-9}  
\multicolumn{2}{c}{}      & ACA$\uparrow$       & Cov.   & Size$\downarrow$ & CCV$\downarrow$ & Cov.   & Size$\downarrow$ & CCV$\downarrow$  \\
\midrule 
\multirow{2}{*}{\rotatebox{0}{\textbf{LAC}}} &
 SCP                     & 50.1 & 0.894 & 4.00 & 8.92 & 0.949 & 4.80 & 5.09 \\
& \our SCA-T             & \our \best{55.2}\imp{+5.1}  & \our 0.898 & \our \best{3.30}\imp{-0.70} & \our \best{7.47}\imp{-1.45}  & \our 0.952 & \our \best{4.03}\imp{-0.77} & \our \best{4.03}\imp{-1.06} \\
\midrule 
\multirow{2}{*}{\rotatebox{0}{\textbf{APS}}} &
SCP                      & 50.1 & 0.896 & 4.05 & 8.70 & 0.953 & 4.87 & 4.82 \\
& \our SCA-T             & \our \best{55.2}\imp{+5.1}  & \our 0.900 & \our \best{3.35}\imp{-0.70} & \our \best{7.18}\imp{-1.52}  & \our 0.954 & \our \best{4.08}\imp{-0.79} & \our \best{3.97}\imp{-0.85} \\
\midrule 
\multirow{2}{*}{\rotatebox{0}{\textbf{RAPS}}} &
SCP                      & 50.1 & 0.895 & 4.15 & 8.75 & 0.953 & 5.01 & 4.90 \\
& \our SCA-T             & \our \best{55.2}\imp{+5.1}  & \our 0.903 & \our \best{3.40}\imp{-0.75} & \our \best{7.50}\imp{-1.25}  & \our 0.954 & \our \best{4.18}\imp{-0.83} & \our \best{4.57}\imp{-0.33} \\
\bottomrule
\end{tabular}
\end{table}

\begin{figure*}[h!]
    \begin{center}
        \begin{tabular}{ccc}

         \includegraphics[width=.32\linewidth]{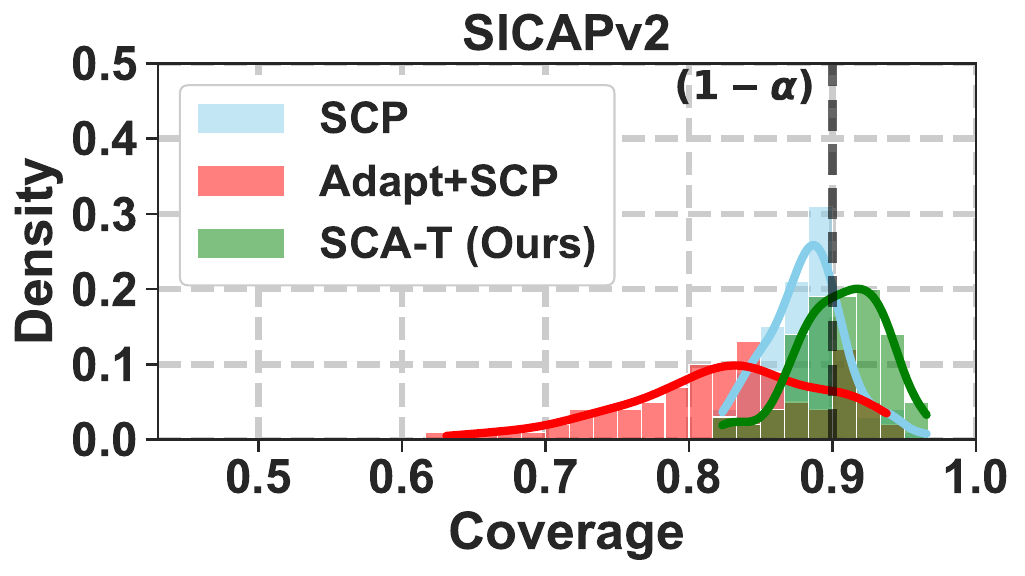} &
         \includegraphics[width=.32\linewidth]{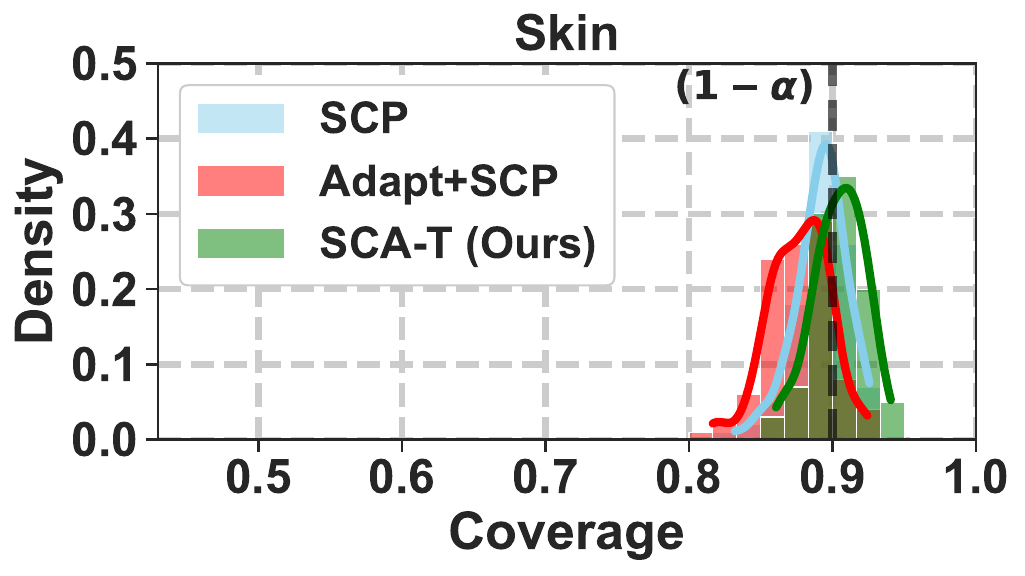} &
         \includegraphics[width=.32\linewidth]{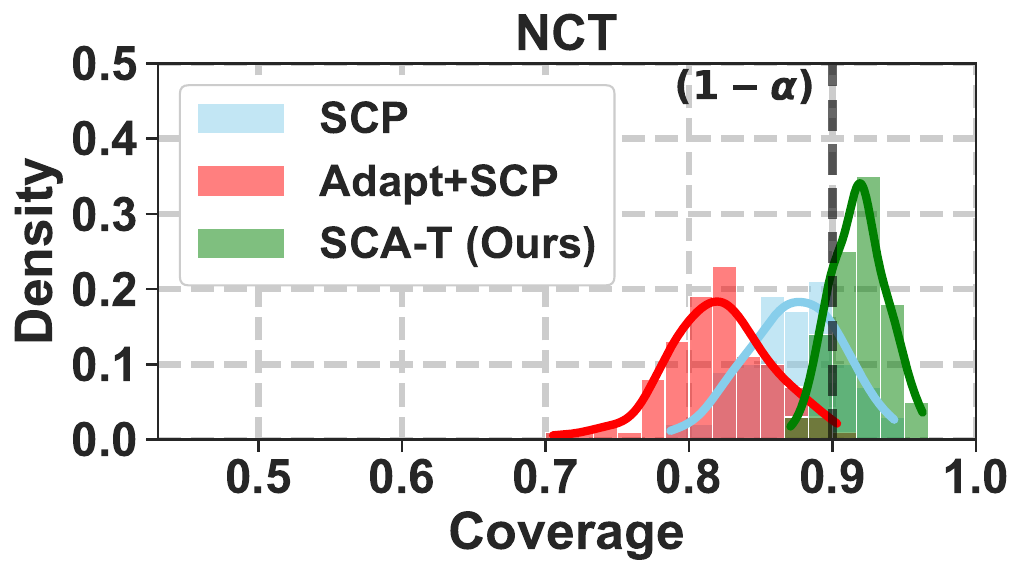} \\

         \includegraphics[width=.32\linewidth]{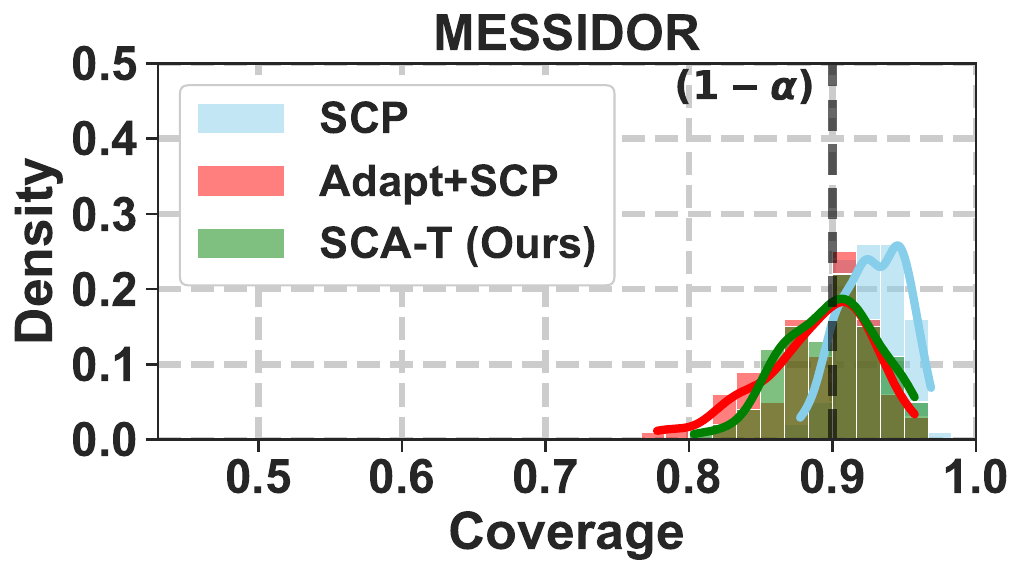} &
         \includegraphics[width=.32\linewidth]{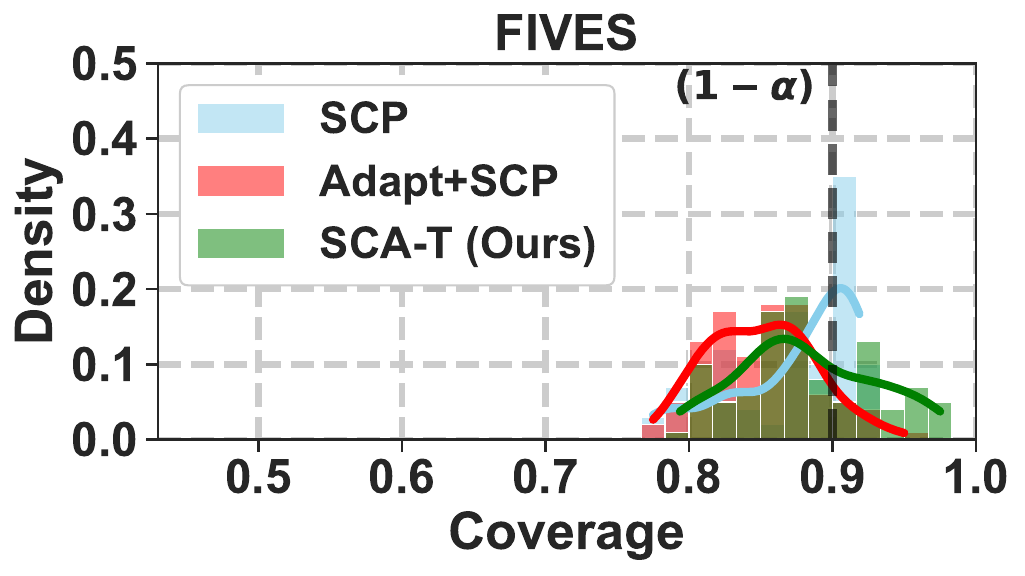} &
         \includegraphics[width=.32\linewidth]{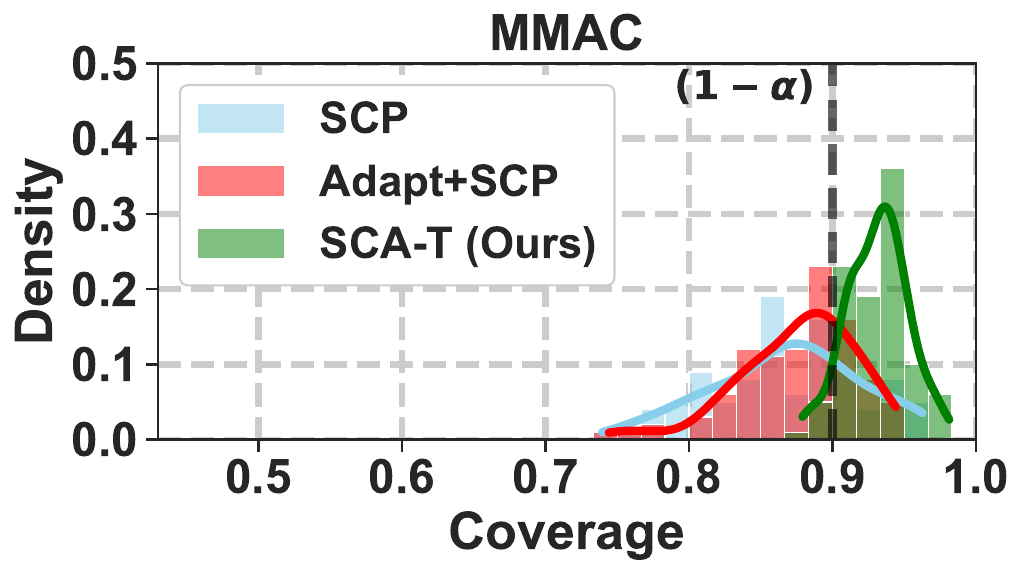} \\

         \includegraphics[width=.32\linewidth]{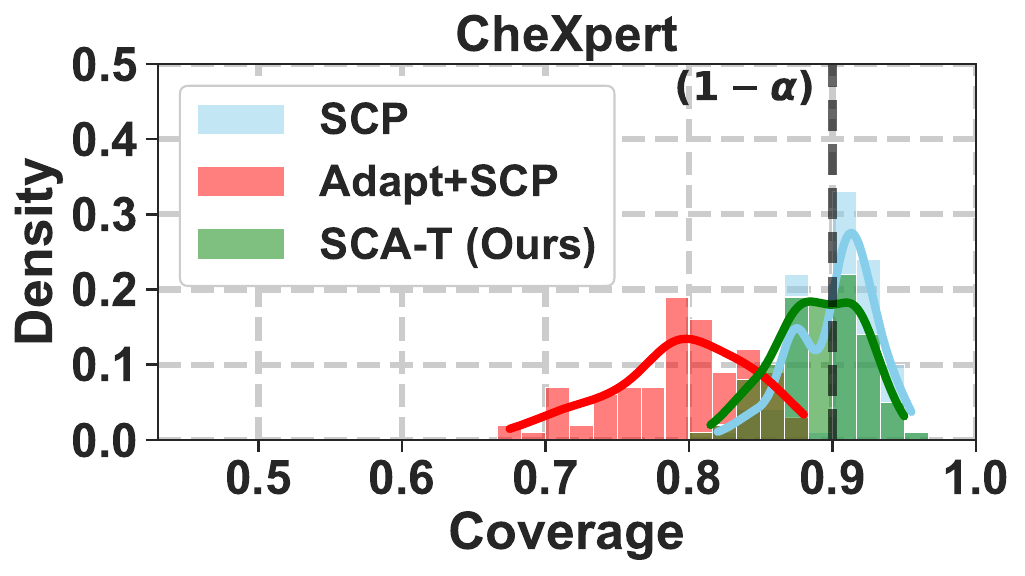} &
         \includegraphics[width=.32\linewidth]{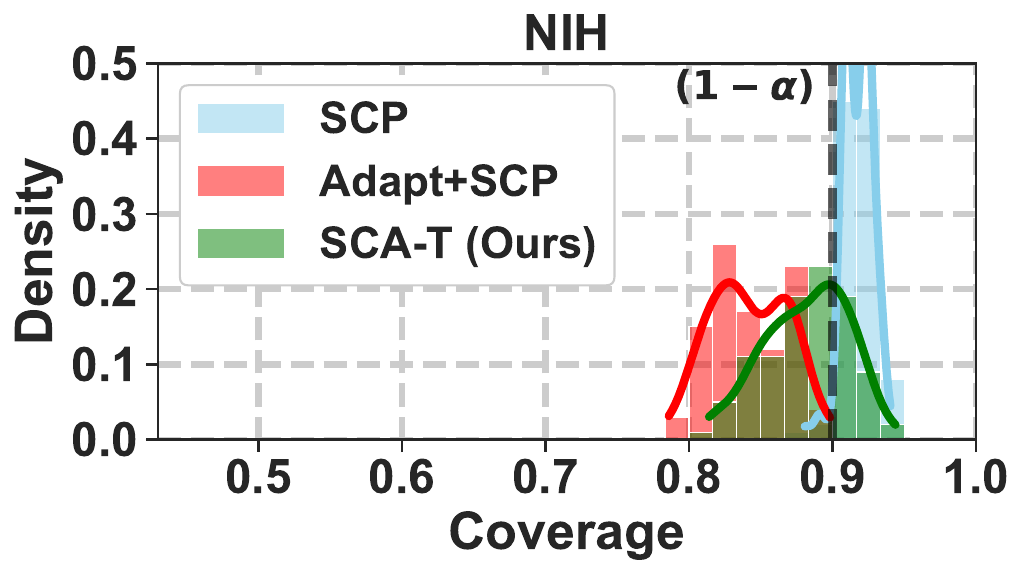} &
         \includegraphics[width=.32\linewidth]{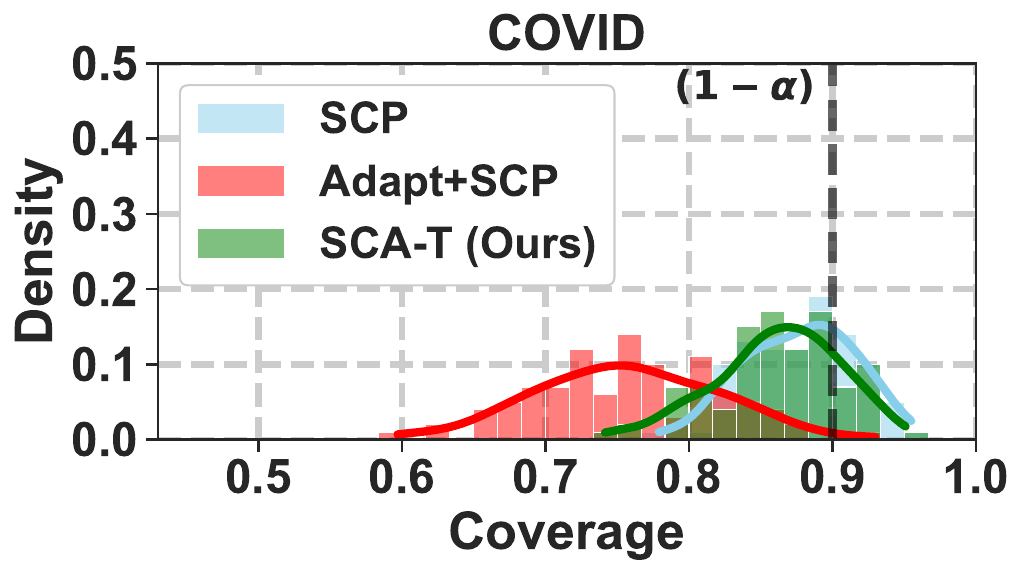}

        \end{tabular}
        \caption{\textbf{Coverage analysis} per dataset and setting using LAC $(\alpha=0.10)$.}
        \label{fig:coverage_datasets}
    \end{center}
\end{figure*}

\subsection{Results}

\noindent\textbf{\textit{Main results.}} Table \ref{main_results} includes the CP results averaged across tasks. These demonstrate the effectiveness of the SCA-T. The proposed framework improves set efficiency by nearly $18\%$ and adaptability (conditional coverage) by $7\%-18\%$. The improvements are consistent across all non-conformity scores and error rates. It is worth noting that combining SCA-T with adaptive scores (e.g., APS) is particularly promising, especially if considering conditional coverage.

\begin{table}[!ht]
\caption{\textbf{Transductive solvers} ($\alpha=0.10$). \textcolor{granate}{\textbf{Red}:} \textcolor{granate}{unsatisfied error rates}.}
\label{tab:transductive_baselines}
\centering
\begin{tabular}{lccccccc}
\toprule
\multicolumn{1}{c}{\multirow{1}{*}{Method}}      & ACA$\uparrow$  & T$\downarrow$ & GPU$\downarrow$ & Cov.   & Size$\downarrow$ & CCV$\downarrow$\\
\midrule 
LAC & 50.1 & \best{0.00} & - & 0.894 & 4.00 & 8.92 \\ \hdashline
\hspace{1mm}TIM \cite{tim} & 53.5\imp{+3.4} & 1.12 & 0.6 & \nc{0.888} & 3.96\imp{-0.04} & 8.08\imp{-0.84} \\
\hspace{1mm}TransCLIP \cite{zanella2024boosting} & 54.8\imp{+4.7} & 0.47 & 1.1 & \nc{0.726} & \best{2.16}\imp{-1.84} & 22.31\wor{+13.39} \\
\our \hspace{1mm}\textit{Ours} & \our \best{55.2}\imp{+5.1}  & \our 1.04 & \our 0.6  & \our 0.898 & \our {3.30}\imp{-0.70} & \our \best{7.47}\imp{-1.45} \\
\midrule 
APS & 50.1 & \best{0.00} & - & 0.896 & 4.05 & 8.69 \\ \hdashline
\hspace{1mm}TIM \cite{tim} & 53.5\imp{+3.4} & 1.18 & 0.6 & \nc{0.887} & 3.16\imp{-0.89} & 7.47\imp{-1.22} \\
\hspace{1mm}TransCLIP \cite{zanella2024boosting} & 54.8\imp{+4.7} & 0.40 & 1.1 & \nc{0.733} & \best{2.52}\imp{-1.53} & 21.78\wor{+13.09} \\
\our \hspace{1mm}\textit{Ours} & \our \best{55.2}\imp{+5.1}  & \our 1.15 & \our 0.6  & \our 0.900 & \our {3.35}\imp{-0.70} & \our \best{7.18}\imp{-1.51} \\
\bottomrule
\end{tabular}
\end{table}

\begin{figure*}[htb]
    \begin{center}
            \begin{tabular}{cccc}
            \includegraphics[width=.243\linewidth]{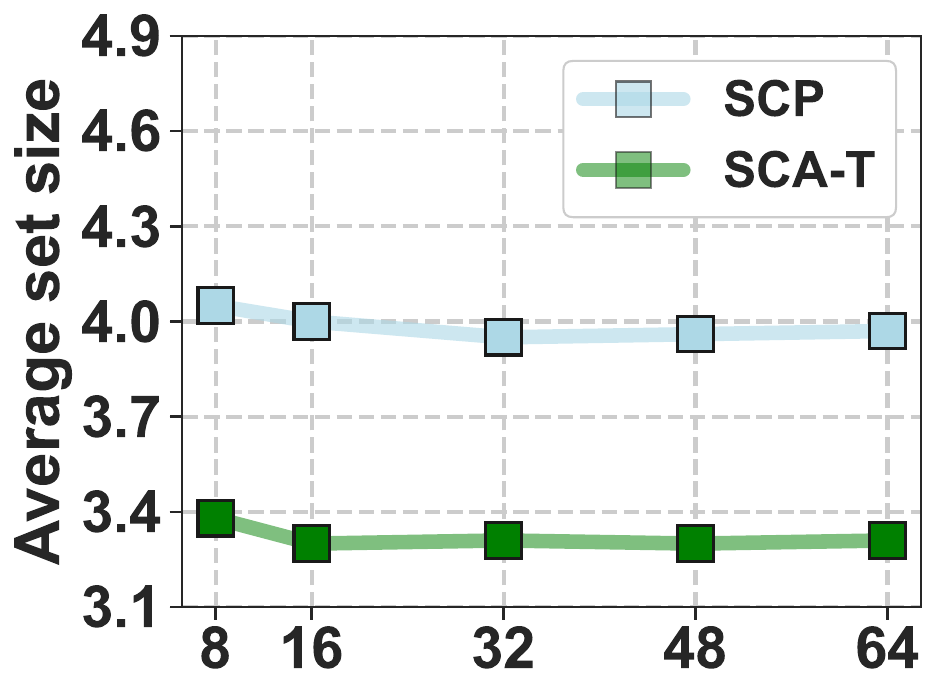} &
            \includegraphics[width=.243\linewidth]{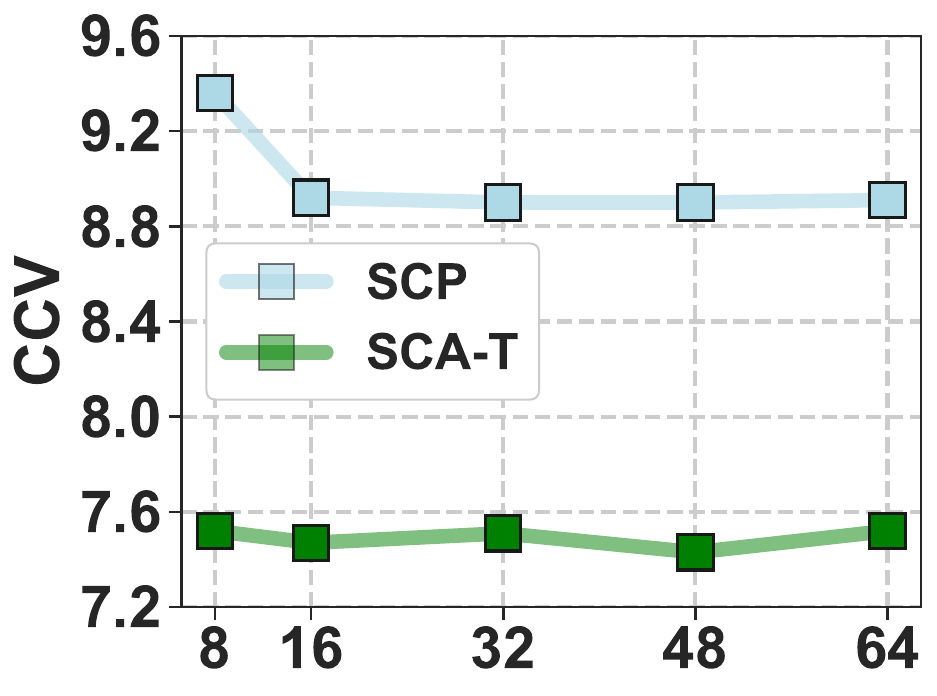} &
            \includegraphics[width=.243\linewidth]{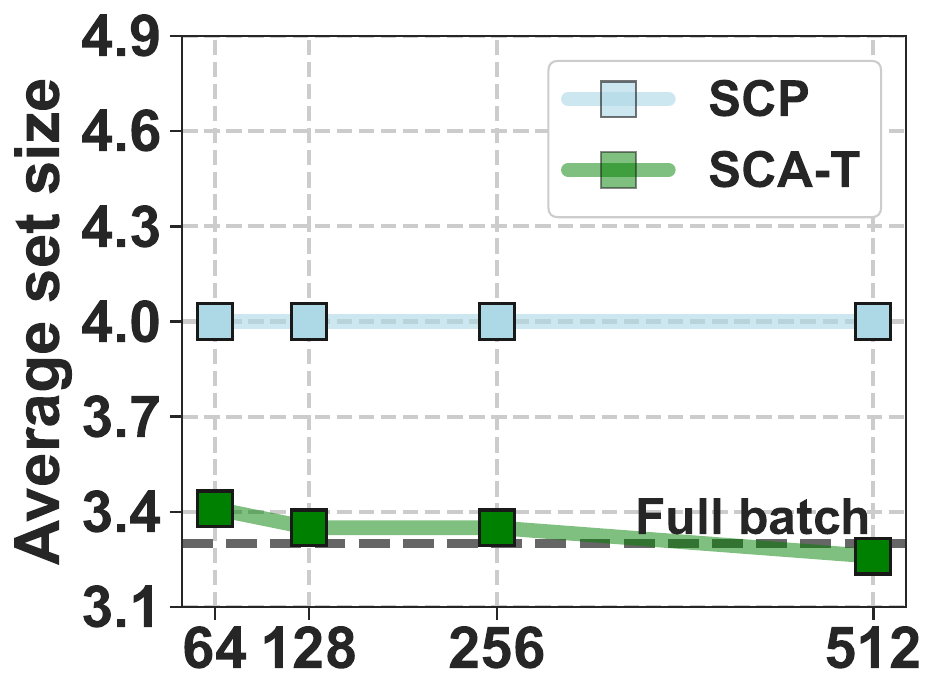} &
            \includegraphics[width=.243\linewidth]{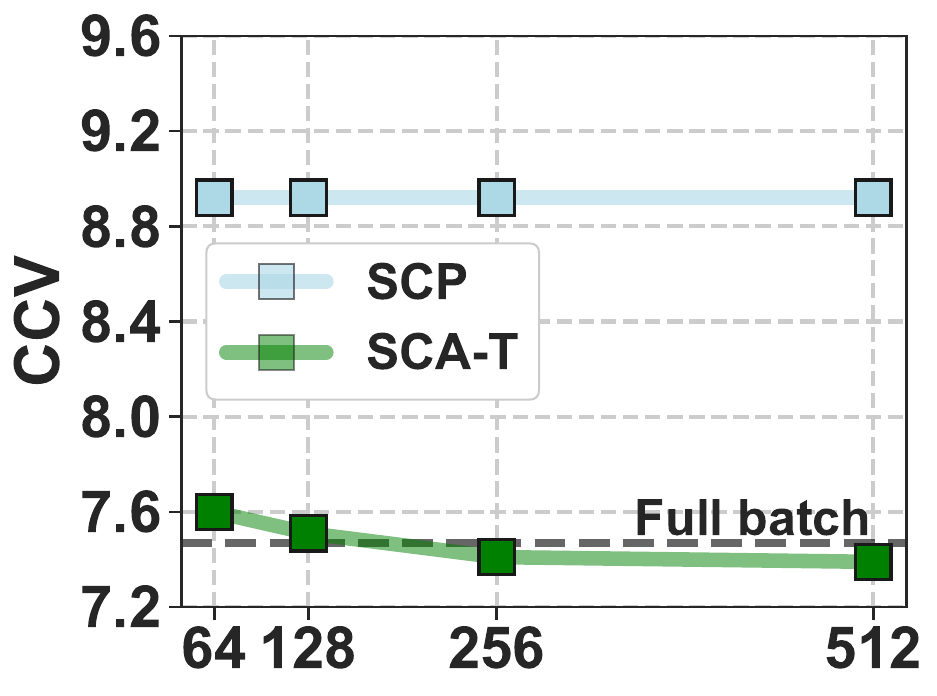} \\
            \multicolumn{2}{c}{\scriptsize{\textbf{(a) Calibration data (K)}}} & \multicolumn{2}{c}{\scriptsize{\textbf{(b) Query set sizes}}}
             \end{tabular}
        \caption{\textbf{Detailed studies.} Results using LAC $(\alpha=0.10)$, averaged across tasks.}
        \label{fig:data_efficiency}
    \end{center}
\end{figure*}

\noindent\textbf{\textit{Coverage analysis.}} Fig. \ref{fig:coverage_datasets} shows the coverage distribution across random seed trials for each task and conformal strategy. One can notice that adapting and conforming the VLM outputs with the same calibration set (i.e., Adapt+SCP) breaks the exchangability of cal/test, and produces suboptimal coverage guarantees. In contrast, SCP provides nearly satisfactory coverage in most scenarios, particularly considering finite-sample error. Furthermore, the proposed SCA-T consistently provides at least the same empirical coverage guarantees as SCP, often exceeding them, as in the NCT and MMAC datasets.

\noindent\textbf{\textit{Baselines.}} Table \ref{tab:transductive_baselines} compares the proposed solver w.r.t. transductive methods. First, note that our solver provides the best accuracy increase (+5.1). Second, the baselines fail to meet the desired coverage, which showcases the importance of regularizing the marginal data distributions during transfer learning. Hence, even though TransCLIP produces smaller sets, those are unreliable. This evaluation also highlights the computational efficiency of SCA-T, which requires only an additional 0.8 seconds for inference and minimal computing resources.

\noindent\textbf{\textit{Detailed analysis.}} Fig. \ref{fig:data_efficiency} explores the limits of SCA-T on challenging data regimes. Results demonstrate consistent improvements over SCP for (a) small calibration sets, even in low-shot settings such as $K=8$, and (b) processing test data in sequentially small test batches instead of operating fully transductive.

 \section{Discussion}
\label{sec:conclusion}

We have explored split conformal prediction in medical VLMs as a tool for trustworthy image classification, and SCA-T as an efficient variant to improve the properties of the conformal sets. Despite its positive performance, some limitations remain. First, the performance of the entropy minimization-based solver in SCA-T relies on the quality of the initial VLM zero-shot predictions, which is less flexible than our previous full-conformal adaptation scenario, FCA \cite{fca25}. In scenarios where compute is not a limiting factor or when addressing only a few target categories, FCA could be a more appealing alternative. Second, CP provides exact theoretical guarantees under exchangeability, which can sometimes be challenging to achieve in healthcare. However, we observe that transductive solvers can mitigate such issues and warrant further research in the future.

\begin{credits}
\subsubsection{\ackname} This work was funded by the Natural Sciences and Engineering Research Council of Canada (NSERC). We also thank Calcul Quebec and Compute Canada.
\end{credits}

\bibliographystyle{splncs04}
\bibliography{refs}

\end{document}